# UN-NORMALIZED HYPERGRAPH P-LAPLACIAN BASED SEMI-SUPERVISED LEARNING METHODS

Loc Tran
Laboratoire CHArt EA4004
EPHE-PSL University, France
tran0398@umn.edu

Linh Tran
Ho Chi Minh University of Technology, Vietnam
linhtran.ut@gmail.com

**Abstract:** Most network-based machine learning methods assume that the labels of two adjacent samples in the network are likely to be the same. However, assuming the pairwise relationship between samples is not complete. The information a group of samples that shows very similar pattern and tends to have similar labels is missed. The natural way overcoming the information loss of the above assumption is to represent the dataset as the hypergraph. Thus, in this paper, we will present the un-normalized hypergraph p-Laplacian semi-supervised learning methods. These methods will be applied to the zoo dataset and the tiny version of 20 newsgroups dataset. Experiment results show that the accuracy performance measures of these un-normalized hypergraph p-Laplacian based semi-supervised learning methods are significantly greater than the accuracy performance measure of the un-normalized hypergraph Laplacian based semi-supervised learning method (the current state of the art hypergraph Laplacian based semi-supervised learning method for classification problem with *p=2*).

1. **Introduction**

To classify the samples is the important problem in machine learning research area. Identifying the class of samples by human effort is very expensive and hard. Hence a lot of computational methods have been proposed to infer the classes of the samples.

To predict the class of the sample, graph which is the natural model of relationship between samples can also be employed. In this model, the nodes represent the samples and the edges represent for the possible interactions between nodes. Then, machine learning methods such as neighbor counting method [1,2], Artificial Neural Networks [3,4], Support Vector Machine [5,6], the un-normalized graph Laplacian based semi-supervised learning method [7,8, 16, 17, 18, 19], or the symmetric normalized and random walk graph Laplacian based semi-supervised learning methods [9,10] can be applied to this graph to infer the classes of un-labeled samples. While the nearest-neighbor classifiers are un-supervised learning methods, the Artificial Neural Networks and the Support Vector Machine are supervised learning methods. The un-normalized, random walk, and symmetric normalized graph Laplacian based semi-supervised learning methods are semi-supervised learning methods.

The Artificial Neural Networks, the SVM method, and three graph Laplacian based semi-supervised learning methods can apply to both single network and multiple networks to infer the classes of the samples. For multiple networks, the SVM method and three graph Laplacian based semi-supervised learning methods try to use weighted combination of these networks (i.e. kernels). [6] (SVM method) determines the optimal weighted combination of networks by solving the semi-definite problem. [7] (un-normalized graph Laplacian based semi-supervised learning method) uses a dual problem and gradient descent to determine the weighted combination of networks. [11] uses the integrated network combined with equal weights, i.e. without optimization due to the integrated network combined with optimized weights has similar performance to the integrated network combined with equal weights and the high time complexity of optimization methods.

The un-normalized, symmetric normalized, and random walk graph Laplacian based semi-supervised learning methods are developed based on the assumption that the labels of two adjacent samples in the network are likely to be the same [7,8,9,10]. Hence this assumption can be interpreted as pair of samples showing a similar pattern and thus sharing edges in the network tends to have similar classes.

In detail, in 2002, Xiaojin Zhu et al. developed the random walk graph Laplacian based semi-supervised learning method and applied this method to various classical applications such as digit recognition [9]. In 2004, Dengyong Zhou et al. developed the symmetric normalized graph Laplacian based semi-supervised learning method and applied this method to various classical applications such as digit recognition and text classification [10]. In 2005 and 2009, Koji Tsuda et al. developed the un-normalized graph Laplacian based semi-supervised learning method and applied this method successfully to the practical bio-informatics problem which is the protein function prediction problem.

In [12,13,14,15], the feature dataset is used for classification problem. However, assuming the pairwise relationship between samples is not complete; the information a group of samples that shows very similar pattern and tends to have similar classes [12,13,14,15] is missed. The natural way overcoming the information loss of the above assumption is to represent the feature dataset as the hypergraph [12,13,14,15]. A hypergraph is a graph in which an edge (i.e. a hyper-edge) can connect more than two vertices. In [12,13], Dengyong Zhou et al. developed the symmetric normalized hypergraph Laplacian based semi-supervised learning method and successfully applied this method to various applications such as digit recognition and text classification. In [14,15], the un-normalized and random walk hypergraph Laplacian based semi-supervised learning methods, which is the two variants of the method developed by Dengyong Zhou, have been developed and successfully applied to protein function prediction and speech recognition problems by Loc Tran in 2014. In general, these three hypergraph Laplacian based semi-supervised learning methods successfully outperform the un-normalized, symmetric normalized, and random walk graph Laplacian based semi-supervised learning methods in classification problem and are completely studied.

To the best of our knowledge, the un-normalized hypergraph p-Laplacian based semi-supervised learning methods have not yet been developed and obviously have not been applied to any practical applications. This method is worth investigated because its nature is difficult and because of its close connection to partial differential equation on hypergraph field. Specifically, in this paper, the un-normalized hypergraph p-Laplacian based semi-supervised learning methods will be developed based on the un-normalized hypergraph p-Laplacian operator definition such as the curvature operator of hypergraph (i.e. the un-normalized hypergraph 1-Laplacian operator). Then these un-normalized hypergraph p-Laplacian based semi-supervised learning methods will be applied to the zoo dataset and the tiny version of 20 newsgroups dataset. Please note that the un-normalized hypergraph p-Laplacian based semi-supervised learning method is the generalization of the un-normalized hypergraph Laplacian based semi-supervised learning method.

We will organize the paper as follows: Section 2 will introduce the preliminary notations and definitions used in this paper. Section 3 will introduce the definition of the gradient and divergence operators of hypergraphs. Section 4 will introduce the definition of Laplace operator of hypergraphs and its properties. Section 5 will introduce the definition of the curvature operator of graphs and its properties. Section 6 will introduce the definition of the p-Laplace operator of graphs and its properties. Section 7 will show how to derive the algorithm of the un-normalized hypergraph p-Laplacian based semi-supervised learning method from regularization framework. In section 8, we will compare the accuracy performance measures of the un-normalized hypergraph Laplacian based semi-supervised learning algorithm (i.e. the current state of art method) and the un-normalized hypergraph p-Laplacian based semi-supervised learning algorithms. Section 9 will conclude this paper and the future direction of researches utilizing discrete operator of hypergraph will be discussed.

2. **Preliminary notations and definitions**

Given a hypergraph $G=(V,E)$, where $V$ is the set of vertices and $E$ is the set of hyper-edges. Each hyper-edge $e \in E$ is the subset of $V$. Please note that the cardinality of $e$ is greater than or equal two. In the other words, $|e| \geq 2$, for every $e \in E$. Let $w(e)$ be the weight of the hyper-edge $e$. Then $W$ will be the $R^{|E|*|E|}$ diagonal matrix containing the weights of all hyper-edges in its diagonal entries.

The incidence matrix $H$ of $G$ is a $R^{|V|*|E|}$ matrix that can be defined as follows

$$h(v,e) = \begin{cases} 1 \text{ if vertex } v \text{ belongs to hyperedge } e \\ 0 \text{ otherwise} \end{cases} \quad (1)$$

From the above definition, we can define the degree of vertex $v$ and the degree of hyper-edge $e$ as follows

$$d(v) = \sum_{e \in E} w(e) * h(v,e) \quad (2)$$

$$d(e) = \sum_{v \in V} h(v,e) \quad (3)$$

Let $D_v$ and $D_e$ be two diagonal matrices containing the degrees of vertices and the degrees of hyper-edges in their diagonal entries respectively. Please note that $D_v$ is the $R^{|V|*|V|}$ matrix and $D_e$ is the $R^{|E|*|E|}$ matrix.

Please note that, we assume that the weight of each hyper-edge is 1.

The inner product on the function space $R^V$ is

$$<f,g>_V = \sum_{u \in V} f_u g_u \quad (4)$$

Also define an inner product on the space of functions $R^E$ on the edges

$$<F,G>_E = \sum_e \sum_{(u,v) \subset e} F_{uv} G_{uv} \quad (5)$$

Here let $H(V) = (R^V, <.,.>_V)$ and $H(E) = (R^E, <.,.>_E)$ be the Hilbert space real-valued functions defined on the vertices of the hypergraph $G$ and the Hilbert space of real-valued functions defined in the edges of $G$ respectively.

3. **Gradient and Divergence Operators**

We define the gradient operator $d: H(V) \to H(E)$ to be (for each hyper-edge $e$)

$$(df)_{uv} = \sqrt{\frac{w(e)}{d(e)}} h(u,e) h(v,e) (f_v - f_u), \quad (6)$$

where $f: V \to R$ be a function of $H(V)$.

We define the divergence operator $div: H(E) \to H(V)$ to be

$$<df, F>_{H(E)} = <f, -div F>_{H(V)}, \quad (7)$$

where $f \in H(V), F \in H(E)$

Next, we need to prove that

$$(div F)_v = \sum_{e \in E} \sum_u \sqrt{\frac{w(e)}{d(e)}} h(u,e) h(v,e) (F_{vu} - F_{uv})$$

Proof:

$$< df, F > = \sum_e \sum_{(u,v) \subset e} df_{uv} F_{uv}$$

$$= \sum_e \sum_{(u,v) \subset e} \sqrt{\frac{w(e)}{d(e)}} h(u,e) h(v,e) f_v F_{uv} - \sum_e \sum_{(u,v) \subset e} \sqrt{\frac{w(e)}{d(e)}} h(u,e) h(v,e) f_u F_{uv}$$

$$= \sum_{k \in V} \sum_e \sum_u \sqrt{\frac{w(e)}{d(e)}} h(u,e) h(k,e) f_k F_{uk} - \sum_{k \in V} \sum_e \sum_v \sqrt{\frac{w(e)}{d(e)}} h(k,e) h(v,e) f_k F_{kv}$$

$$= \sum_{k \in V} \sum_e \sum_u \sqrt{\frac{w(e)}{d(e)}} h(u,e) h(k,e) f_k F_{uk} - \sum_{k \in V} \sum_e \sum_u \sqrt{\frac{w(e)}{d(e)}} h(k,e) h(u,e) f_k F_{ku}$$

$$= \sum_{k \in V} f_k \sum_{e \in E} \sum_u \sqrt{\frac{w(e)}{d(e)}} h(u,e) h(k,e) (F_{uk} - F_{ku})$$

Thus, we have

$$(div F)_v = \sum_{e \in E} \sum_u \sqrt{\frac{w(e)}{d(e)}} h(u,e) h(v,e) (F_{vu} - F_{uv}) \ (8)$$

4. **Laplace operator**

We define the Laplace operator $\Delta: H(V) \to H(V)$ to be

$$\Delta f = -\frac{1}{2} div(df) \ (9)$$

Next, we compute

$$(\Delta f)_v = \frac{1}{2} \sum_{e \in E} \sum_u \sqrt{\frac{w(e)}{d(e)}} h(u,e) h(v,e) (df_{uv} - df_{vu})$$

$$= \frac{1}{2} \sum_{e \in E} \sum_u \sqrt{\frac{w(e)}{d(e)}} h(u,e) h(v,e) \left( \sqrt{\frac{w(e)}{d(e)}} h(u,e) h(v,e) (f_v - f_u) - \sqrt{\frac{w(e)}{d(e)}} h(u,e) h(v,e) (f_u - f_v) \right)$$

$$= \sum_{e \in E} \sum_u \frac{w(e)}{d(e)} h(u,e) h(v,e) (f_v - f_u)$$

$$= \sum_{e \in E} \sum_u \frac{w(e)}{d(e)} h(u,e) h(v,e) f_v - \sum_{e \in E} \sum_u \frac{w(e)}{d(e)} h(u,e) h(v,e) f_u$$

$$= f_v \sum_{e \in E} \frac{w(e)}{d(e)} h(v,e) \sum_u h(u,e) - \sum_{e \in E} \sum_u \frac{w(e)}{d(e)} h(u,e) h(v,e) f_u$$

$$= f_v \sum_{e \in E} \frac{w(e)}{d(e)} h(v,e) d(e) - \sum_{e \in E} \sum_u \frac{w(e)}{d(e)} h(u,e) h(v,e) f_u$$

$$= f_v \sum_{e \in E} w(e)h(v,e) - \sum_{e \in E} \sum_u \frac{w(e)}{d(e)} h(u,e)h(v,e)f_u$$

$$= f_v d(v) - \sum_{e \in E} \sum_u \frac{w(e)}{d(e)} h(u,e)h(v,e)f_u$$

Thus, we have

$$(\Delta f)_v = f_v d(v) - \sum_{e \in E} \sum_u \frac{w(e)}{d(e)} h(u,e)h(v,e)f_u \quad (10)$$

The hypergraph Laplacian is a linear operator. Furthermore, the hypergraph Laplacian is self-adjoint and positive semi-definite.

Let $S_2(f) = \frac{1}{2}\sum_i \|d_i f\|^2$, we have the following **theorem 1**

$$D_f S_2 = \Delta f \quad (11)$$

5. **Curvature operator**

We define the curvature operator $\kappa: H(V) \to H(V)$ to be

$$\kappa f = -\frac{1}{2} div\left(\frac{df}{\|df\|}\right) \quad (12)$$

Next, we compute

$$(\kappa f)_v = \frac{1}{2} \sum_{e \in E} \sum_u \sqrt{\frac{w(e)}{d(e)}} h(u,e)h(v,e)\left(\left(\frac{df}{\|df\|}\right)_{uv} - \left(\frac{df}{\|df\|}\right)_{vu}\right)$$

$$= \frac{1}{2} \sum_{e \in E} \sum_u \sqrt{\frac{w(e)}{d(e)}} h(u,e)h(v,e)\left(\frac{1}{\|d_u f\|}\sqrt{\frac{w(e)}{d(e)}} h(u,e)h(v,e)(f_v - f_u) - \frac{1}{\|d_v f\|}\sqrt{\frac{w(e)}{d(e)}} h(u,e)h(v,e)(f_u - f_v)\right)$$

$$= \frac{1}{2} \sum_{e \in E} \sum_u \frac{w(e)}{d(e)} h(u,e)h(v,e)\left(\frac{1}{\|d_u f\|} + \frac{1}{\|d_v f\|}\right)(f_v - f_u)$$

Thus, we have

$$(\kappa f)_v = \frac{1}{2} \sum_{e \in E} \sum_u \frac{w(e)}{d(e)} h(u,e)h(v,e)\left(\frac{1}{\|d_u f\|} + \frac{1}{\|d_v f\|}\right)(f_v - f_u) \quad (13)$$

From the above formula, we have

$$d_u f = ((df)_{uv}: (u,v) \subset e, \forall e)^T \quad (14)$$

The local variation of $f$ at $u$ is defined to be

$$\|d_u f\| = \sqrt{\sum_e \sum_{(u,v) \subset e}(df)_{uv}^2} = \sqrt{\sum_e \sum_{(u,v) \subset e} \frac{w(e)}{d(e)} h(u,e)h(v,e)(f_v - f_u)^2} \quad (15)$$

To avoid the zero denominators in (11), the local variation of $f$ at $u$ is defined to be

$$\|d_u f\| = \sqrt{\sum_e \sum_{(u,v) \subset e}(df)^2_{uv} + \epsilon}, \quad (16)$$

where $\epsilon = 10^{-10}$.

The hypergraph curvature is a non-linear operator.

Let $S_1(f) = \sum_u \|d_u f\|$, we have the following **theorem 2**

$$D_f S_1 = \kappa f \quad (17)$$

6. **p-Laplace operator**
   We define the p-Laplace operator $\Delta_p : H(V) \to H(V)$ to be

$$\Delta_p f = -\frac{1}{2} div(\|df\|^{p-2} df) \quad (18)$$

Clearly, $\Delta_1 = \kappa$ and $\Delta_2 = \Delta$. Next, we compute

$$(\Delta_p f)_v = \frac{1}{2} \sum_{e \in E} \sum_u \sqrt{\frac{w(e)}{d(e)}} h(u,e) h(v,e) (\|df\|^{p-2} df_{uv} - \|df\|^{p-2} df_{vu})$$

$$= \frac{1}{2} \sum_{e \in E} \sum_u \sqrt{\frac{w(e)}{d(e)}} h(u,e) h(v,e) (\|d_u f\|^{p-2} \sqrt{\frac{w(e)}{d(e)}} h(u,e) h(v,e)(f_v - f_u) - \|d_v f\|^{p-2} \sqrt{\frac{w(e)}{d(e)}} h(u,e) h(v,e)(f_u - f_v))$$

$$= \frac{1}{2} \sum_{e \in E} \sum_u \frac{w(e)}{d(e)} h(u,e) h(v,e)(\|d_u f\|^{p-2} + \|d_v f\|^{p-2})(f_v - f_u)$$

Thus, we have

$$(\Delta_p f)_v = \frac{1}{2} \sum_{e \in E} \sum_u \frac{w(e)}{d(e)} h(u,e) h(v,e)(\|d_u f\|^{p-2} + \|d_v f\|^{p-2})(f_v - f_u) \quad (19)$$

Let $S_p(f) = \frac{1}{p} \sum_i \|d_i f\|^p$, we have the following **theorem 3**

$$D_f S_p = \Delta_p f \quad (20)$$

7. **Discrete regularization on hypergraphs and classification problems**
   Given a hypergraph $G=(V,E)$, where $V$ is the set of vertices and $E$ is the set of hyper-edges. Each hyper-edge $e \in E$ is the subset of $V$. Let $y$ denote the initial function in $H(V)$. $y_i$ can be defined as follows

$$y_i = \begin{cases} 1 & if\ sample\ i\ belongs\ to\ the\ class \\ -1 & if\ sample\ i\ does\ not\ belong\ to\ the\ class \\ 0 & otherwise \end{cases}$$

Our goal is to look for an estimated function $f$ in $H(V)$ such that $f$ is not only smooth on $G$ but also close enough to an initial function $y$. Then each sample $i$ is classified as $sign(f_i)$. This concept can be formulated as the following optimization problem

$$argmin_{f \in H(V)} \{S_p(f) + \frac{\mu}{2} \|f - y\|^2\} \quad (21)$$

The first term in (21) is the smoothness term. The second term is the fitting term. A positive parameter $\mu$ captures the trade-off between these two competing terms.

### 7.1) 2-smoothness

When $p=2$, the optimization problem (21) is

$$argmin_{f \in H(V)}\{\frac{1}{2}\sum_i \|d_i f\|^2 + \frac{\mu}{2}\|f - y\|^2\} \quad (22)$$

By theorem 1, we have

**Theorem 4:** The solution of (22) satisfies

$$\Delta f + \mu(f - y) = 0 \quad (23)$$

Since $\Delta$ is a linear operator, the closed form solution of (23) is

$$f = \mu(\Delta + \mu I)^{-1} y, \quad (24)$$

Where $I$ is the identity operator and $\Delta = D_v - HWD_e^{-1}H^T$. (24) is the algorithm proposed by [14,15].

### 7.2) 1-smoothness

When $p=1$, the optimization problem (21) is

$$argmin_{f \in H(V)}\{\sum_i \|d_i f\| + \frac{\mu}{2}\|f - y\|^2\}, \quad (25)$$

By theorem 2, we have

**Theorem 5:** The solution of (25) satisfies

$$\kappa f + \mu(f - y) = 0, \quad (26)$$

The curvature $\kappa$ is a non-linear operator; hence we do not have the closed form solution of equation (25). Thus, we have to construct iterative algorithm to obtain the solution. From (26), we have

$$\frac{1}{2}\sum_{e \in E}\sum_u \frac{w(e)}{d(e)} h(u,e) h(v,e) (\frac{1}{\|d_u f\|} + \frac{1}{\|d_v f\|})(f_v - f_u) + \mu(f_v - y_v) = 0 \quad (27)$$

Define the function $m: E \to R$ by

$$m_{uv} = \frac{1}{2}\sum_{e \in E} \frac{w(e)}{d(e)} h(u,e) h(v,e) (\frac{1}{\|d_u f\|} + \frac{1}{\|d_v f\|}) \quad (28)$$

Then equation (27) which is

$$\sum_u m_{uv}(f_v - f_u) + \mu(f_v - y_v) = 0$$

can be transformed into

$$(\sum_u m_{uv} + \mu) f_v = \sum_u m_{uv} f_u + \mu y_v \quad (29)$$

Define the function $p: E \to R$ by

$$p_{uv} = \begin{cases} \frac{m_{uv}}{\Sigma_u m_{uv}+\mu} & if\ u \neq v \\ \frac{\mu}{\Sigma_u m_{uv}+\mu} & if\ u = v \end{cases} \quad (30)$$

Then

$$f_v = \Sigma_u p_{uv} f_u + p_{vv} y_v\ (31)$$

Thus, we can consider the iteration

$$f_v^{(t+1)} = \Sigma_u p_{uv}^{(t)} f_u^{(t)} + p_{vv}^{(t)} y_v \text{ for all } v \in V$$

to obtain the solution of (25).

### 7.3) p-smoothness

For any number $p$, the optimization problem (21) is

$$argmin_{f \in H(V)}\{\frac{1}{p}\Sigma_i \|d_i f\|^p + \frac{\mu}{2}\|f - y\|^2\},\ (32)$$

By theorem 3, we have

**Theorem 6:** The solution of (32) satisfies

$$\Delta_p f + \mu(f - y) = 0,\ (33)$$

The *p-Laplace* operator is a non-linear operator; hence we do not have the closed form solution of equation (33). Thus, we have to construct iterative algorithm to obtain the solution. From (33), we have

$$\frac{1}{2}\Sigma_{e \in E}\Sigma_u \frac{w(e)}{d(e)} h(u,e) h(v,e)(\|d_u f\|^{p-2} + \|d_v f\|^{p-2})(f_v - f_u) + \mu(f_v - y_v) = 0\ (32)$$

Define the function $m: E \to R$ by

$$m_{uv} = \frac{1}{2}\Sigma_{e \in E} \frac{w(e)}{d(e)} h(u,e) h(v,e)(\|d_u f\|^{p-2} + \|d_v f\|^{p-2})\ (33)$$

Then equation (32) which is

$$\sum_u m_{uv}(f_v - f_u) + \mu(f_v - y_v) = 0$$

can be transformed into

$$(\Sigma_u m_{uv} + \mu) f_v = \Sigma_u m_{uv} f_u + \mu y_v\ (34)$$

Define the function $p: E \to R$ by

$$p_{uv} = \begin{cases} \frac{m_{uv}}{\Sigma_u m_{uv}+\mu} & if\ u \neq v \\ \frac{\mu}{\Sigma_u m_{uv}+\mu} & if\ u = v \end{cases} \quad (35)$$

Then

$$f_v = \sum_u p_{uv} f_u + p_{vv} y_v \quad (36)$$

Thus, we can consider the iteration

$$f_v^{(t+1)} = \sum_u p_{uv}^{(t)} f_u^{(t)} + p_{vv}^{(t)} y_v \text{ for all } v \in V$$

to obtain the solution of (32).

## 8. Experiments and results

### Datasets

In this paper, we used the zoo dataset and the tiny version of 20 newsgroups dataset which can be obtained from UCI repository and from http://www.cs.nyu.edu/~roweis/data.html respectively. The zoo data set contains 101 animals with 17 attributes. The attributes include hair, feathers, eggs, milk, etc. The animals have been classified into 7 different classes. In this dataset, each attribute is the hyper-edge. The tiny version of the 20 newsgroups dataset contains the binary occurrence data for 100 words across 16242 postings. However, we just choose small subset of this tiny dataset containing 200 postings in order to test our algorithms. In this dataset, each word is the hyper-edge. Thus, these two datasets are themselves the hypergraphs and we don't need to preprocess these two datasets. These two input datasets are very similar to the following figure 1 of [12,13].

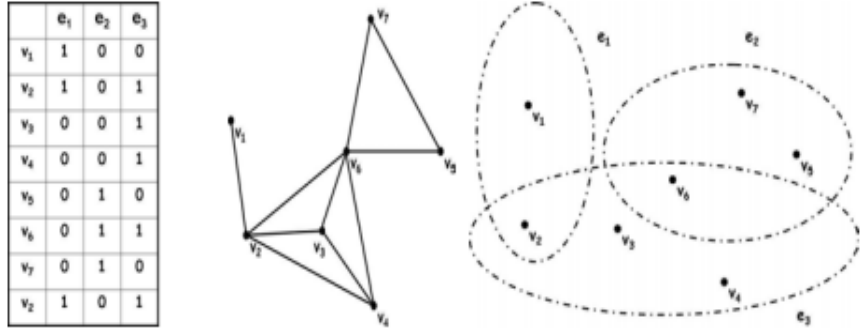

Figure 1. Hypergraph example with 8 vertices and 3 hyper-edges [12,13]

### Experiments and Results

In this section, we experiment with the above proposed un-normalized hypergraph p-Laplacian based semi-supervised learning methods with *p=3, 3.1, 3.2, 3.3, 3.4, 3.5, 3.6, 3.7, 3.8, 3.9, 4.0* and the current state of the art method (i.e. the un-normalized hypergraph Laplacian based semi-supervised learning method *p=2*) applied directly to the zoo dataset and the tiny version of 20 newsgroups dataset in terms of accuracy performance measure. The accuracy performance measure Q is given as follows

$$Q = \frac{True\ Positive + True\ Negative}{True\ Positive + True\ Negative + False\ Positive + False\ Negative}$$

All experiments were implemented in MATLAB 6.5 on virtual machine. The accuracy performance measures of the above proposed methods are given in the following table 1 and table 2.

Table 1: **Accuracies** of the un-normalized hypergraph p-Laplacian based semi-supervised learning methods and the current state of the art hypergraph Laplacian based semi-supervised learning method *p=2* for the **zoo dataset**

| p values | Accuracy performance measures (%) |
|---|---|
| 2 | 85.71 |
| 3 | 88.80 |
| 3.1 | 90.48 |
| 3.2 | 91.04 |
| 3.3 | 91.60 |
| 3.4 | 92.16 |
| 3.5 | 92.44 |
| 3.6 | 92.44 |
| 3.7 | 92.16 |
| 3.8 | 90.20 |
| 3.9 | 89.64 |
| 4.0 | 88.52 |

The following figure 2 shows the **accuracies** of the un-normalized hypergraph p-Laplacian based semi-supervised learning methods and the current state of the art hypergraph Laplacian based semi-supervised learning method *p=2* for the **zoo dataset**:

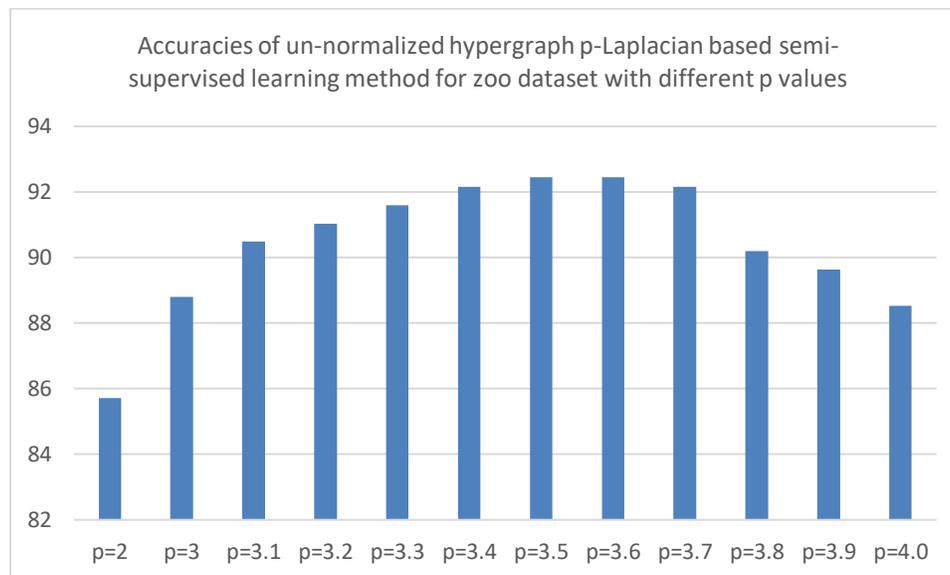

Table 2: **Accuracies** of the un-normalized hypergraph p-Laplacian based semi-supervised learning methods and the current state of the art hypergraph Laplacian based semi-supervised learning method *p=2* for the tiny version of **20 newsgroups dataset**

| p values | Accuracy performance measures (%) |
|---|---|
| 2 | 75.50 |
| 3 | 78.25 |
| 3.1 | 80.00 |
| 3.2 | 81.25 |
| 3.3 | 81.75 |
| 3.4 | 82.25 |
| 3.5 | 82.00 |
| 3.6 | 82.50 |
| 3.7 | 83.50 |
| 3.8 | 83.75 |
| 3.9 | 82.75 |
| 4.0 | 82.75 |

The following figure 2 shows the **accuracies** of the un-normalized hypergraph p-Laplacian based semi-supervised learning methods and the current state of the art hypergraph Laplacian based semi-supervised learning method $p=2$ for the tiny version of the **20 newsgroups dataset**:

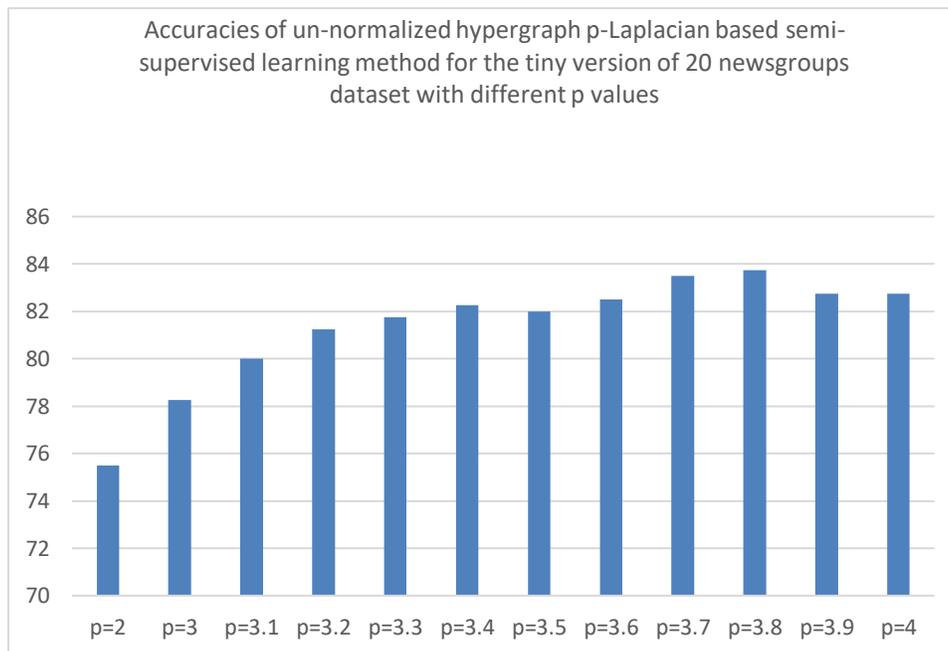

From the above tables, we easily recognized that the un-normalized hypergraph p-Laplacian based semi-supervised learning methods outperform the current state of art method. The results from the above tables show that the un-normalized hypergraph p-Laplacian based semi-supervised learning methods are at least as good as the current state of the art method ($p=2$) but often lead to significant better classification accuracy performance measures.

9. **Conclusions**

In this paper, we have proposed the detailed algorithms of the un-normalized hypergraph p-Laplacian based semi-supervised learning methods applied to the zoo dataset and the tiny version of 20 newsgroups dataset. Interestingly, experiments show that the un-normalized hypergraph p-Laplacian based semi-supervised learning methods are at least as good as the un-normalized hypergraph Laplacian based semi-supervised learning method (the current state of the art method $p=2$) but often lead to significant better classification accuracy performance measures.

To the best of our knowledge, the un-normalized hypergraph p-Laplacian Eigenmaps and the un-normalized hypergraph p-Laplacian clustering methods have not yet been developed. These methods are worth investigated because of their difficult nature and their close connection to partial differential equation on hypergraph field. In the future, we will develop the un-normalized hypergraph p-Laplacian Eigenmaps and the un-normalized hypergraph p-Laplacian clustering methods by constructing the hypergraph p-Laplacian matrix. Then from this hypergraph p-Laplacian matrix, we can also construct another version of the un-normalized hypergraph p-Laplacian based semi-supervised learning methods.